\documentclass{article}
\usepackage[preprint]{log_2022}

\usepackage{booktabs}            
\usepackage{multirow}            
\usepackage{amsfonts}            
\usepackage{graphicx}            
\usepackage{duckuments}          
\usepackage{todonotes}
\usepackage{colortbl}
\usepackage{xcolor}

\usepackage{subcaption}
\usepackage{enumitem}
\usepackage[numbers,compress,sort]{natbib}

\newcommand{\ie}{\textit{i}.\textit{e}.\ }
\newcommand{\eg}{\textit{e}.\textit{g}.\ }

\title[Towards Temporal Edge Regression]{Towards Temporal Edge Regression: A Case Study on Agriculture Trade Between Nations}

\author[]{
Lekang Jiang\textsuperscript{1}\thanks{Equal contribution.} \quad Caiqi Zhang\textsuperscript{1}\footnotemark[1] \quad Farimah Poursafaei\textsuperscript{2,3} \quad Shenyang Huang\textsuperscript{2,3}\\
\{ \email{lj408,cz391}\} \email{@cam.ac.uk} \quad \email{farimah.poursafaei@mila.quebec} \quad \email{shenyang.huang@mail.mcgill.ca}  \\
\textsuperscript{1}University of Cambridge, 
\textsuperscript{2}McGill University, 
\textsuperscript{3}Mila - Quebec AI Institute
}

\begin{document}

\maketitle

\begin{abstract}

Recently, Graph Neural Networks~(GNNs) have shown promising performance in tasks on dynamic graphs such as node classification, link prediction and graph regression. However, few work has studied the temporal edge regression task which has important real-world applications. In this paper, we explore the application of GNNs to edge regression tasks in both static and dynamic settings, focusing on predicting food and agriculture trade values between nations. We introduce three simple yet strong baselines and comprehensively evaluate one static and three dynamic GNN models using the UN Trade dataset. Our experimental results reveal that the baselines exhibit remarkably strong performance across various settings, highlighting the inadequacy of existing GNNs. We also find that TGN outperforms other GNN models, suggesting TGN is a more appropriate choice for edge regression tasks. Moreover, we note that the proportion of negative edges in the training samples significantly affects the test performance. The companion source code can be found at: \url{https://github.com/scylj1/GNN_Edge_Regression}.

\end{abstract}

\section{Introduction}

Graph representation learning has gained significant attention in recent years due to its ability to model complex relationships and structures in various domains \citep{kazemi2020representation}. Graph Neural Networks (GNNs), as a core technique within this field, have shown remarkable success in various applications, ranging from social network analysis \citep{ying2018graph, monti2019fake, frasca2020sign} to the biological field \citep{zitnik2018modeling, li2021braingnn, huang2020edge}. This growing interest has led to the development of various GNN architectures optimized for specific tasks and objectives.

Although most existing GNN models, such as GCN \citep{kipf2016semi}, GAT \citep{velivckovic2017graph}, GraphSAGE \citep{hamilton2017inductive}, and GIN \citep{xu2018powerful}, have been developed for static graphs, there is a growing interest in representation learning on dynamic graphs, where the graph structure changes over time. The temporal graph representation learning is particularly useful in modeling real-life dynamic systems, such as social networks, traffic networks, and biological systems, where the graph topology evolves as time passes \citep{kazemi2020representation}. Therefore, it is crucial to develop effective methods for modeling the dynamics of graph-structured data. Numerous works have focused on both discrete-time dynamic graphs \citep{yu2018netwalk, pareja2020evolvegcn, sankar2020dysat} and continuous-time dynamic graphs \citep{kumar2019predicting, trivedi2019dyrep, xu2020inductive, rossi2020temporal}.


However, there are few attempts on applying GNNs to edge regression tasks for dynamic networks~\cite{sharma2022representation}. Most of the existing works have focused on node classification, link prediction, and graph classification tasks \citep{kazemi2020representation}. In this paper, we address this research gap by exploring the application of both static and dynamic GNNs to edge regression tasks on the UN Trade dataset \cite{poursafaei2022towards}, where food and agriculture trade values between nations are predicted. This real-life problem offers a unique opportunity to investigate the performance of GNNs in terms of edge regression, as the trade relationships between countries are naturally represented as a graph with time-varying edge weights. Results of edge regression on this dataset not only improve the understanding of applying GNNs on edge regression but also assist in forecasting the future global food and agriculture market, which is crucial for policy making and resource allocation.

One of the key observation is that among the GNN models, TGN achieves the best performance, indicating that temporal attention mechanisms and memory modules can better capture the evolving dynamics of the graph. Additionally, we observe that the percentage of negative edges in all training samples has a significant impact on the test performance. Overall, our work represents a step toward developing effective methods for modeling temporal edge regression tasks using GNNs.

Overall, this work has the following contributions:

\begin{itemize}
    \item We propose three straightforward but strong baselines to compare the effectiveness of existing GNN models.

    \item We implement three normalization methods, three training strategies, and two evaluation criteria to optimize and measure the performance of the models.

    \item We conduct a thorough evaluation and analysis of one static and three dynamic GNN models using the UN Trade dataset. Our experiments demonstrate that the baselines achieve surprisingly strong performance across multiple settings, underscoring that existing GNNs are not sufficiently effective.
\end{itemize}

\section{Related Work}

\textbf{Dynamic Graph Representation Learning.} Aiming at learning node representations that evolve over time, dynamic graph representation learning is a rapidly developing field in recent years. Recently, several approaches have been proposed to address the challenge of dynamic graph representation learning, such as JODIE \citep{kumar2019predicting}, DyRep \citep{trivedi2019dyrep}, TGAT \citep{xu2020inductive}, and TGN \citep{rossi2020temporal}. \citet{kazemi2020representation} and \citet{skarding2021foundations} provide detailed surveys of advances in representation learning on dynamic graphs and a detailed terminology of dynamic networks. However, according to the reported experiment results of the recent methods, they often achieve close to perfect performance for current link prediction tasks on dynamic graphs, which hinders researchers' ability to evaluate if new models are superior \citep{poursafaei2022towards,souza2022provably,cong2022we,huang2023temporal}. Therefore, to better compare the strengths and weaknesses of emergent dynamic graph neural networks, \citet{poursafaei2022towards} propose six new dynamic graph datasets in different domains with two more challenging negative sampling strategies. Our work thus focuses on the UN Trade dataset from \citep{poursafaei2022towards} to investigate the existing temporal GNNs' capabilities for edge regression tasks. 

\textbf{Edge Regression Tasks.} According to the survey by~\citet{kazemi2020representation}, the main applications of dynamic graph representation learning include link prediction \citep{kumar2019predicting, trivedi2019dyrep, xu2020inductive, rossi2020temporal, chen2019lstm}, entity/relation prediction \citep{kumar2019predicting, leblay2018deriving, dasgupta2018hyte}, recommender systems \citep{kumar2018learning}, time prediction \citep{dasgupta2018hyte}, node classification \citep{pareja2020evolvegcn, sato2019dyane}, and graph classification \citep{taheri2019learning}. Despite the success of various models in achieving state-of-the-art results on the abovementioned tasks, to the best of our knowledge, there has been little research on the edge regression task. One reason for this is the added complexity of edge regression, which involves predicting a continuous value rather than a binary or categorical label. Additionally, there is few datasets available for the edge regression task, posing challenges in evaluating and comparing different algorithms for edge regression. Thus, our work aims to address the gap in the research on edge regression tasks by utilizing the UN Trade dataset \citep{poursafaei2022towards}.

\section{Methodology}

\subsection{Task Formulation}
\begin{figure}[!ht]
    \centering
    \includegraphics[width=.99\textwidth]{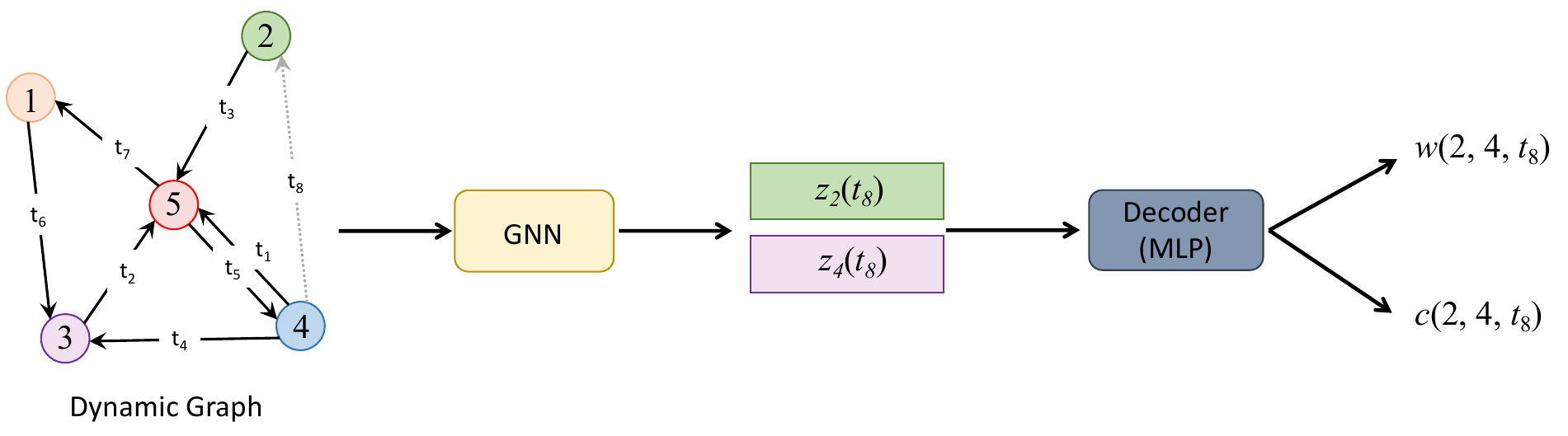}
    \caption{Example of a GNN encoder ingesting a dynamic graph with seven visible edges (with timestamps $t_1$ to $t_7$), with the goal of predicting the future weight $w(2, 4, t_8)$ (or weight class $c(2, 4, t_8)$) between nodes 2 and 4 at time $t_8$. $z_2(t_8)$ and $z_4(t_8)$ are the embeddings for nodes 2 and 4 at time $t_8$.}
    \label{fig:pipeline}
\end{figure}

\textbf{Temporal Edge Regression Task.} Following the definition from~\citep{poursafaei2022towards}, a dynamic graph can be represented as timestamped edge streams - triplets of source, destination, timestamp, i.e. $\mathcal{G}=\left\{\left(s_0, d_0, t_0\right),\left(s_1, d_1, t_1\right), \ldots,\left(s_T, d_T, T\right)\right\}$ where the timestamps are ordered $\left(0 \leq t_1 \leq t_2 \leq \ldots \leq t_{s p l i t} \leq \ldots \leq T\right)$ . We denote the set of all nodes in $\mathcal{G}$ at time $t_n$ by $\mathcal{V}(t_n)$ and let $\mathcal{E}(t_n)$ be the set of all edges in $\mathcal{G}$ at time $t_n$. Consequently, $\mathcal{E}(s_n, t_n)$ is the set of all edges in $\mathcal{E}(t_n)$ that originate from node $s_n$ at time $t_n$. We investigate the task of predicting the exact weight of an edge between a node pair in the future, \ie $w(s_i, d_j, t_n)$. Figure~\ref{fig:pipeline} shows an example of our temporal edge regression task. 

\textbf{Temporal Edge Classification Task.} Due to the inherent difficulty of the edge regression task, an alternative approach to predict the trading value is to convert it to a classification problem. In this approach, the range of edge values is partitioned into $n$ intervals, with each interval corresponding to a class. For instance, the intervals could be defined as follows in the log space: $(0, 1], (1, 10], (10, 10^2], (10^2, 10^3],\ldots,(10^{n-1}, 10^n]$. By transforming the regression task to a magnitude classification task, the difficulty of the prediction task is reduced. The goal of the classification task is now to evaluate how well existing GNNs can estimate the order of magnitude of the edge weights (\ie $c(s_i, d_j, t_n)$), rather than predicting the exact values of edges. 


\subsection{Normalization}

Normalization involves scaling the weights so that they fall within a specific range or have a particular statistical distribution. In the context of the UN Trade dataset, the trade values between countries vary significantly. Thus, normalizing the weights is beneficial for the performance of these models by ensuring that the input data falls within a fixed / expected range. We explore the following three normalization methods in this work:

\textbf{Min-max normalization.} It scales the values of a dataset from their natural range into a standard range (\eg from $a$ to $b$). The formula for min-max normalization can be expressed mathematically as:

\begin{equation}
w^* = a + \frac{(w - \min(x))(b-a)}{\max(w) - \min(w)}
\end{equation}

where $w$ is the original weight, $w_{norm}$ is the normalized value, $\min(w)$ and $\max(w)$ are the minimum and maximum values in the dataset, respectively, and $a$ and $b$ are the lower and upper bounds of the desired range respectively. 

\textbf{Log normalization.} In log normalization, it takes the logarithm of the values in a dataset. The formula for log normalization can be expressed simply as:

\begin{equation}
w^* = \log(w)
\end{equation}

Log normalization is commonly used when the data has a wide range of values or when the distribution of values is highly skewed. By taking the logarithm of the values, the range of values can be compressed, and the distribution can be transformed into a more symmetric shape. 

\textbf{Node degree normalization.} While min-max and log normalization are applied across all timestamps, we introduce a novel normalization method named node degree normalization for each timestamp. For a given timestamp $t_n$ and node $s_n$, the proposed normalization method divides all weights originating from node $s_n$ by their sum. Specifically, this method divides the weight $w(s_n, d_m, t_n)$ of each edge between node $s_n$ and destination node $d_m$ at time $t_n$ by the sum of the raw weights of all edges originating from $s_n$ at time $t_n$. This results in the normalized weight $w_{norm}(s_n, d_m, t_n)$ for each edge, calculated as:

\begin{equation}
w^*(s_n, d_m, t_n) = \frac{w(s_n, d_m, t_n)}{\sum_{d_k \in \mathcal{V}(s_n, t_n)} w(s_n, d_k, t_n)}
\end{equation}

where $\mathcal{V}(s_n, t_n)$ is the set of all nodes at timestamp $t_n$ that originate from $s_n$. This normalization method ensures that the weights of edges originating from node $s_n$ sum up to 1 at any given time, facilitating comparison of the relative strengths of different connections originating from that node. By utilizing node degree normalization, the model does not predict the exact trade values anymore but rather predicts the proportion of the total trade value of a country exporting to another country in following years. 

\subsection{Models}

\textbf{Baselines. } In this study, we implement three baseline methods served as benchmarks to compare with different GNN models and evaluate their effectiveness. 

\begin{itemize}[leftmargin=*]

    \item \textbf{Mean/Most}. The \textbf{Mean} baseline is applied in the edge regression task, which predicts all edges using the average value of all inputs.

    \begin{equation}
    w = \frac{1}{np}\sum_{j=1}^{p}\sum_{i=1}^{n}w_i(t_j)
    \end{equation}
    
    where $t$ is the timestamp, $t_p$ is the last timestamp in the training set, and $n$ is the number of positive edges at a given snapshot. 
    In the classification task, a similar \textbf{Most} baseline is applied to predict the edge weights using the class which appears the most frequently in the training dataset. 
    
    \begin{equation}
    c = \text{mode}(c_{1}, c_{2}, ..., c_{n})
    \end{equation}

    \item \textbf{Persistence Forecast}. It predicts each edge using the last seen value (class) of that edge in the training set. 

    \begin{equation}
    w(i,j , t_{q}) = w(i,j, t_{p})
    \end{equation}
    where $t_q$ is the current timestamp and $t_p$ is the timestamp of the most recent record of the edge between node $i$ and $j$.

    \item \textbf{Historical Average}. It predicts each edge's weight by computing the average value of all the weights observed for that edge in the training set.
    
    \begin{equation}
    w(i,j, t_{q}) = \frac{1}{p}\sum_{k=1}^{p} w(i,j, t_k)
    \end{equation}
\end{itemize}

\begin{figure}[!ht]
    \centering
    \includegraphics[width=.99\textwidth]{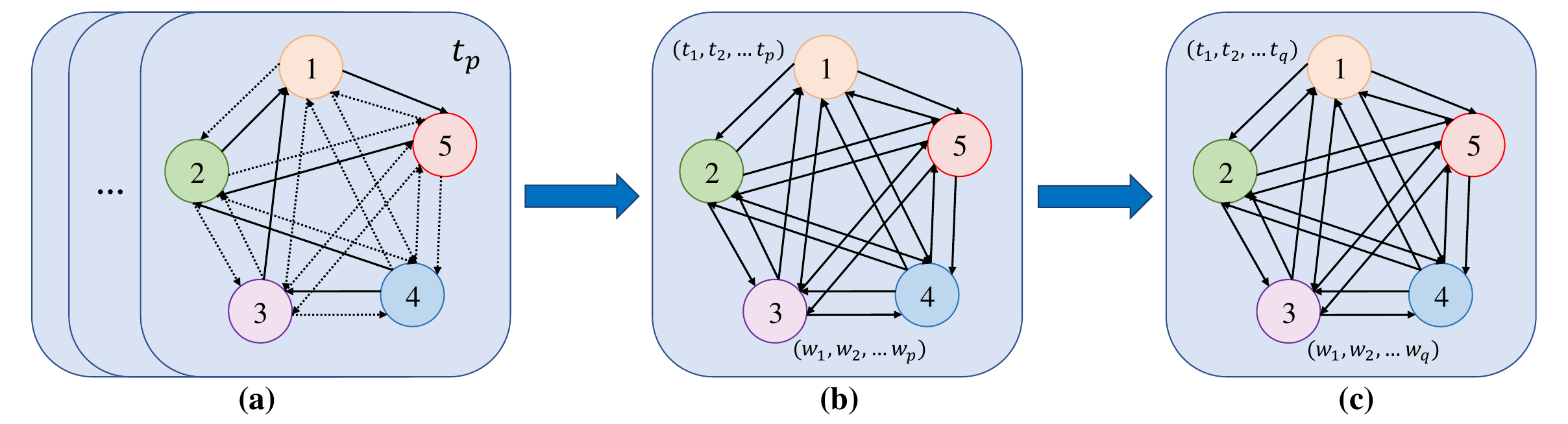}
    \caption{Illustration of transforming temporal graph to static graph for GCN training and testing; (a) fully-connected graph at each timestamp, (b) collapsed static graph with feature vectors of length $p$, (c) adjusted static graph with feature vectors of length $q$. }
    \label{fig:gcn}
\end{figure}

\textbf{Static GNN. } We implement GCN \cite{kipf2016semi} as a representative static GNN model for comparison. As GCN cannot deal with temporal information, we construct collapsed static graphs for training and testing in the following steps. We first develop fully-connected graphs at each timestamp by filling the non-existing edges with the value zero. Second, compressing all training graphs from $t_1$ to $t_p$ into a single static graph, where the node features are a list of timestamps $(t_1, t_2, \ldots, t_p)$ and edge features are a list of edge values at each timestamp $(w_1, w_2, \ldots, w_p)$, as shown in Figure~\ref{fig:gcn}. The same process is also applied to validation and test set. Then, we transform the node features and edge features of length $p$ in the training set to match the length $q$ of the test set using the following equation. We split the feature vectors of length $p$ into $q$ groups equally, and then calculate the average value of each group to obtain the transformed result. After pre-processing, we obtain 3 static graphs that have the same dimensions of features and outputs for training, validation and testing.

\begin{equation}
   w = \left \{   \frac{1}{m} \sum_{j=1}^{m} w_{(i-1)m+j} \mid i \in \left [ 1, p \right ] \right \}
\end{equation}
where $m$ is the number of elements in each group.

\textbf{Dynamic GNNs. } The three dynamic GNNs modified for the task of temporal edge regression are as follows:

\begin{itemize}[leftmargin=*]
\item JODIE \citep{kumar2019predicting}: It is designed for bipartite networks of instantaneous user-item interactions. It consists of an update operation and a projection operation. The update operation utilizes two coupled RNNs to recursively update the representation of the users and items. The projection operation predicts the future representation of a node while considering the elapsed time since its last interaction.

\item DyRep \citep{trivedi2019dyrep}: It utilizes a specialized RNN to update node representations when a new edge is observed. To compute neighbor weights at each time step, DyRep employs a temporal attention mechanism that is parameterized on the recurrent architecture.

\item TGN \citep{rossi2020temporal}: consists of five key modules: (1) Memory, which stores the historical data of each node and facilitates the retention of long-term dependencies; (2) Message function, which updates the memory of each node based on the messages generated when an event is observed; (3) Message aggregator, which combines multiple messages involving a single node; (4) Memory updater, which updates the memory of a node based on the aggregated messages; and (5) Embedding, which generates representations of nodes by considering the node's memory and its associated edge and node features.

\end{itemize}

\section{Experiments}
\subsection{Dataset}

The United Nations food and agriculture trade dataset is originally collected, processed, and disseminated by the Food and Agriculture Organization of the United Nations \citep{poursafaei2022towards}. The data is primarily provided by UNSD, Eurostat, and other national authorities as required. The trade data contains statistics on all food and agriculture products that are imported or exported annually by all countries worldwide. Specifically, in the UN Trade dataset, the graph $\mathcal{G}$ represents a weighted, directed food and agriculture trading graph among 181 nations, spanning from 1986 to 2017, with around 500,000 edges. The edge weights indicate the total sum of normalized agriculture import or export values between two countries. This dataset provides a valuable resource for studying global food and agriculture trade and can be used in a variety of applications, including graph neural network models for predicting future trade patterns. The distribution of edge weight values in UN Trade dataset is illustrated in Appendix~\ref{a:data} (Figure~\ref{fig:distribution}).

\subsection{Training Strategies}
\begin{figure}[!ht]
    \centering
    \includegraphics[width=.99\textwidth]{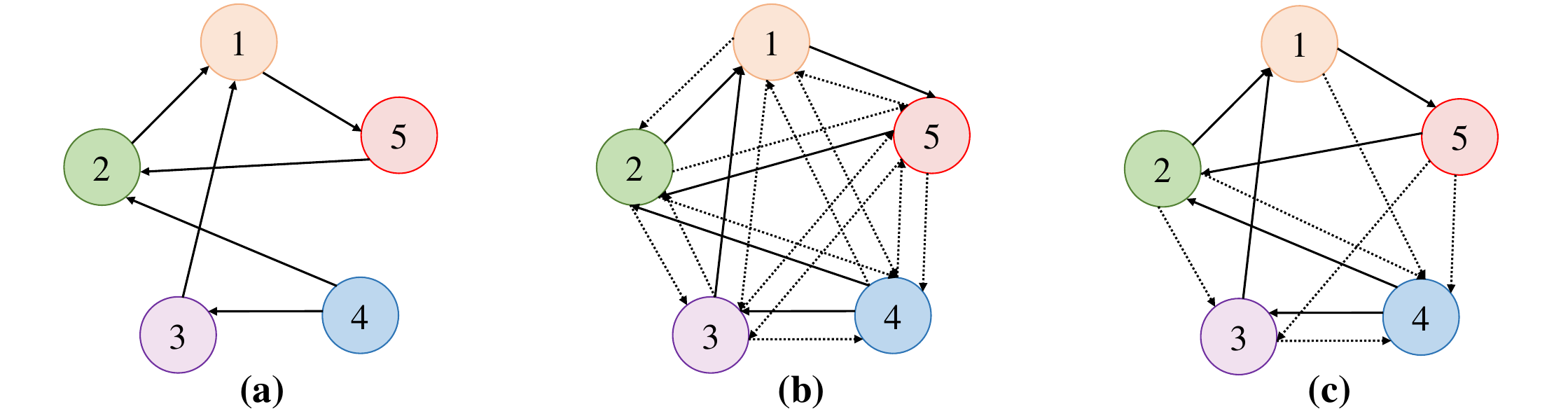}
    \caption{Three training strategies. Solid lines represent edges with positive values that actually exist in the data set (positive edges). Dashed lines represent virtual edges with a weight of zero (negative edges). (a) train on positive edges, (b) train on all edges, (c) train with negative sampling. }
    \label{fig:train}
\end{figure}

As few previous works have focused on edge regression tasks, it is unclear how to train models to achieve optimal results. Thus, we utilize the following three training strategies to compare their performance, which are demonstrated in Figure \ref{fig:train} respectively.

\textbf{Training with negative sampling.} Negative sampling refers to randomly selecting a negative edge for each positive edge during training. It reduces the computational complexity of training while still allowing the model to learn from negative samples. This is the standard training strategies used for link prediction task for GNNs~\cite{rossi2020temporal, huang2023temporal, kazemi2020representation, trivedi2019dyrep}. 

\textbf{Training on positive edges.} We refer positive edges to those edges that indeed exist between two nodes (\ie with positive edge values). A negative edge is a node pair where an edge hasn't been observed in the temporal graph.  
The advantage of training solely on positive edges is that it simplifies and accelerates the training process, and the model can better capture the features on positive edges to make precise predictions.


\textbf{Training on all node pairs.} To train on all node pairs, we construct a fully-connected graph for each snapshot, where each positive edge is of its original value, and negative edges are set to zero. The advantage of this training strategy is that the model can learn from the full range of relationships presented in the network. A potential risk is that this approach can be computationally expensive~(with complexity $O(|\mathcal{V}(t_n)|^2)$ ) and memory-intensive, particularly for large networks.

\subsection{Evaluation Strategies}

\textbf{Evaluation with negative sampling.} For all positive edges in the test set, we randomly sample the same number of negative edges for evaluation, which is consistent with past papers \cite{rossi2020temporal, poursafaei2022towards}. Evaluating on both positive and negative edges can assess the model's overall performance, and we refer this type of evaluation to the \textbf{Overall} evaluation. 

\textbf{Evaluation on positive edges.} We argue that in real-world edge regression applications, predicting negative edges is sometimes not as crucial as predicting positive edges. For example, people may be more concerned about the actual trade value between two countries rather than focusing on countries with no trade history. Therefore, we propose the \textbf{Positive} evaluation to focus on the more important and relevant edges in the graph, which is closer to real-life scenarios.

\textbf{Old and new nodes.} In temporal graphs, new nodes may emerge over time, like in social networks. Following \citep{rossi2020temporal}'s experiments, we evaluate edges between both old and new nodes (unseen in the training set). Although the emergence of a new nation is rare in our case, we simulate this situation to provide a more comprehensive evaluation of the edge regression task.

\section{Results}

\subsection{Temporal Edge Regression Task}

In this section, we discuss the experiment results on the temporal edge regression task. The predicted MSE losses on edges for old nodes are listed in Table~\ref{t1}. As the creation of new nations is rare, we report the results involving new nodes in Appendix~\ref{a1} (Table~\ref{t2}). The observations and findings of new nodes are similar to old nodes. 

\begin{table}[!t]
\caption{Loss with standard deviation (old nodes). Numbers in \textcolor{red}{red} indicate the best results for all methods, and numbers in \textbf{bold} are the best results among GNNs. }
\centering
\resizebox{\textwidth}{!}{
\begin{tabular}{lcccccc}
\toprule
                                             & \multicolumn{2}{c}{Log normalization}                                       & \multicolumn{2}{c}{\begin{tabular}[c]{@{}c@{}}Min-max normalization \\ ($10^{-2}$) \end{tabular}}                  & \multicolumn{2}{c}{\begin{tabular}[c]{@{}c@{}}Node degree normalization \\ ($10^{-3}$)\end{tabular}}                   \\ \cmidrule(lr){2-3} \cmidrule(lr){4-5} \cmidrule(lr){6-7}
                                             & Positive MSE                & Overall MSE                                      & Positive MSE                 & Overall MSE                          & Positive  MSE                        & Overall MSE                      \\ \midrule
\textbf{Baselines}                            &                              &                                              &                              &                                  &                                      &                              \\
Mean                                         & 2.762                        & 5.504                                        & 3.933                        & 1.979                            & 2.644                                & 1.429                        \\
Persistence Forecast                                   & {\color[HTML]{FF0000} 0.608} & 4.135                                        & {\color[HTML]{FF0000} 0.104} & 2.125                            & {\color[HTML]{FF0000} 1.267}         & {\color[HTML]{FF0000} 0.943} \\
Historical Average                          & 0.833                        & 3.634                                        & 0.959                        & {\color[HTML]{FF0000} 1.360}      & 1.667                                & 1.224                        \\ \midrule
\textbf{Static GNN}                          &                              &                                              &                              &                                  &                                      &                              \\
GCN                                          & 6.774 (0.56)                 & 4.041 (0.41)                                 & 106.1 (39.3)                 & 67.18 (26.1)                     & 1025 (406.)                          & 656.0 (261.)                 \\ \midrule
\textbf{Dynamic GNNs}                         &                              &                                              &                              &                                  &                                      &                              \\
\multicolumn{7}{l}{\textbf{Train on all node pairs}}\\
TGN                                          & 9.014 (0.05)                 & 4.773 (0.12) & 3.974 (0.00)                 & 1.988 (0.00)                     & 2.687 (0.00)                         & 1.344 (0.00)        \\
JODIE                                        & 8.446 (0.18)                 & 4.580 (0.07)                                 & 12.92 (10.6)                 & 364.8 (464.)                     & 41.41 (36.7)                         & 51.75 (28.8)                 \\
DyRep                                        & 8.875 (0.10)                 & 4.777 (0.04)                                 & 3.972 (0.00)                 & 1.988 (0.00)                     & \cellcolor[HTML]{FFFFFF}2.600 (0.00) & 1.360 (0.00)                 \\ \midrule
\multicolumn{7}{l}{\textbf{Train on positive edges}}                                        \\
TGN                                          & \textbf{1.810 (0.05)}        & 4.241 (0.20)                                 & \textbf{3.960 (0.00)}        & 1.983 (0.00)           & \textbf{2.479 (0.03)}                & 1.355 (0.03)                 \\
JODIE                                        & 2.808 (0.11)                 & 24.09 (22.3)                                 & 8.250 (5.98)                 & 910.1 (128.)                     & 7.519 (4.06)                         & 169.5 (105.)                 \\
DyRep                                        & 3.061 (0.22)                 & 3.831 (0.16)                                 & 3.961 (0.01)                 & 1.994 (0.01)                     & 2.894 (0.19)                         & 2.002 (0.30)                 \\ \midrule
\multicolumn{7}{l}{\textbf{Train with negative sampling}}        \\ 
TGN                                          & 3.802 (0.09)        & {\color[HTML]{FF0000} \textbf{2.959 (0.03)}} & 3.965 (0.00)       & \textbf{1.982 (0.00)}            & 2.522 (0.07)               & \textbf{1.342 (0.05)}        \\
JODIE                                        & 5.009 (0.11)                 & 4.815 (0.33)                                 & 4.157 (0.19)                 & \multicolumn{1}{l}{31.48 (25.2)} & 11.87 (3.62)                         & 29.07 (5.38)                 \\
DyRep                                        & 5.292 (0.21)                 & 3.637 (0.07)                                 & 4.022 (0.07)                 & 2.115 (0.09)                     & \cellcolor[HTML]{FFFFFF}2.832 (0.01) & 3.234 (0.79)                 \\ \bottomrule
\end{tabular}}
\label{t1}
\end{table}

\textbf{Comparison with baselines. }The results show that the Persistence Forecast baseline exhibits the best performance for most cases (4/6), and the Historical Averages baseline reaches the best outcome of 1.360 on Overall MSE loss using min-max normalization (1/6). This observation reveals that our baselines are strong and existing GNNs are not yet able to outperform the baselines, highlighting the research gap in edge regression tasks. There are several reasons for this result: first, the food and agriculture trade values between the countries are normally stable over time, leading to a strong performance of the Persistence Forecast baseline. Second, the test set only contains a relatively short period of time (4 years), and significant value fluctuations are unlikely to occur within this period. Third, the training data (22 timestamps) may be insufficient for complex models to acquire a full understanding of the underlying patterns. 

\textbf{Comparisons among GNNs.} Regarding the performance of GNN models, TGN trained with negative sampling demonstrates the best Overall MSE loss of 2.959 compared to all the other models. This may be because TGN trained with an appropriate number of negative edges, can better differentiate negative edges, resulting in lower Overall MSE loss. It suggests that TGN is a feasible approach for temporal edge regression tasks as it can learn temporal dependencies and graph structure to improve predictions. Another possible reason is that TGN employs temporal attention mechanisms and specialized memory modules, which capture the evolving relationships and dynamics of the network over time more effectively. This may result in more accurate predictions and a deeper insight of the underlying temporal patterns.  

Notably, although JODIE is effective in predicting future interactions \cite{kumar2019predicting}, it may not be suitable for edge regression tasks because its performance is much worse and more unstable than other models in some situations. For example, it results in the Overall MSE loss of 910.1 with a standard deviation of 128 when training on positive edges. It is worth mentioning that dynamic GNNs typically exhibit superior performance compared to static GCN because of their ability to leverage the temporal characteristics of the graph effectively. 

\textbf{Influence of training strategies.} Our findings show that no GNN model trained on all edges achieves optimal results on Positive and Overall MSE loss. We hypothesize that this is due to the sparsity of the graph, making it challenging for models to learn meaningful representations from training on all edges, resulting in worse performance. In contrast, we observe that TGN trained on positive edges achieves the best Positve MSE loss in all three normalization methods, and TGN trained with negative sampling demonstrates the best Overall MSE loss. 

Furthermore, we note a trade-off between the Overall MSE loss and Positive MSE loss when different numbers of negative edges engaged in training. Training with more negative edges leads to a lower Overall MSE loss and a higher Positive MSE loss. For instance, when using log normalization, training with negative sampling can reduce the Overall MSE loss from 4.241 to 2.959, while increasing the Positive MSE loss from 1.810 to 3.802. One future research direction is to determine the optimal percentage of negative edges used during training to achieve a balanced regression performance between positive and negative edges.

\textbf{Influence of normalization methods.} Although both log and min-max normalization scale the original edge values to a standard range to facilitate fast and effective training of GNNs, we argue that log normalization is more suitable for training GNNs for two reasons. Firstly, the loss values using log normalization are more sensible, as GCN and JODIE perform poorly and unstably when using min-max normalization, producing loss values that are hundreds of times larger than the baselines. Secondly, TGN outperforms the baselines in terms of Overall MSE loss only when using log normalization. A possible explanation is that log normalization can create a more uniform data distribution with less skewness and fewer outliers, facilitating effective training on edge regression tasks. 

Unlike log and min-max normalization, node degree normalization enables the model to predict a percentage value of the trade flow between countries, rather than an absolute value. The purpose of this novel normalization approach is to demonstrate that specialized normalization based on graphs can offer another perspective on the dataset. We encourage researchers to explore other normalization methods that could improve the model's performance in edge regression tasks.

\subsection{Temporal Edge Classification Task}

\begin{table}[!t]
    \centering
    \caption{Accuracy and F1 score (\%) with standard deviation (old nodes). Numbers in \textcolor{red}{red} indicate the best results for all methods, and numbers in \textbf{bold} are the best results among GNNs.}
    \begin{tabular}{lcccc}
\toprule
   &  Positive Accuracy            & Overall Accuracy                                 & Positive F1                  & Overall F1                        \\ \midrule
\textbf{Baselines}         &                              &                                              &                              &                              \\
Most                       & 22.14                        & 11.07                                        & 8.21                         & 2.27                         \\
Persistence forecast                & {\color[HTML]{FF0000} 60.97} & 50.00                                           & {\color[HTML]{FF0000} 62.31} & {\color[HTML]{FF0000} 50.68}                      \\ \midrule
\textbf{Dynamic GNNs}      &                              &                                              &                              &                              \\
\multicolumn{5}{l}{\textbf{Train on postive edges}}                               \\
TGN                        & \textbf{32.19 (0.85)}        & 16.10 (0.43)                                 & \textbf{29.95 (1.04)}        & 10.53 (0.40)                 \\
JODIE                      & 22.03 (1.32)                 & 11.01 (0.66)                                 & 14.02 (2.73)                 & 4.61 (1.12)                  \\
DYREP                      & 23.04 (0.96)                 & 11.52 (0.48)                                 & 16.20 (2.17)                 & 5.28 (0.84)                  \\ \midrule
\multicolumn{5}{l}{\textbf{Train with negative sampling}}                              \\
TGN                        & 12.91 (0.98)                 & {\color[HTML]{FF0000} \textbf{50.37 (0.24)}} & 14.81 (0.81)                 & \textbf{43.01 (0.51)}        \\
JODIE                      & 0.38 (0.01)                  & 49.47 (0.67)                                 & 0.33 (0.01)                  & 33.39 (0.33)                 \\
DYREP                      & 0.07 (0.08)                  & 50.01 (0.01)                                 & 0.04 (0.04)                  & 33.38 (0.06)                 \\ \bottomrule
\end{tabular}
\label{t3}
\end{table}

Table~\ref{t3} reports the accuracy and F1 scores of the temporal edge classification task involving only old nodes. The classification results containing new nodes are reported in Appendix~\ref{a1} (Table~\ref{t4}), which is similar to the results with old nodes. In this task, we focus only on dynamic GNNs trained on positive edges or with negative sampling, because we find that in the regression task, dynamic GNNs outperform static GCN, and training on all edges often leads to unsatisfactory results. Note that historical average is not applicable in this classification task. 

Compared to the regression task, the classification task is inherently easier, and the evaluation metrics (\ie accuracy and F1 score) are also more intuitive. The Persistence Forecast baseline remains powerful, outperforming GNNs in Positive accuracy (60.97\%), Positive F1 (62.31\%), and Overall F1 score (50.68\%). Moreover, TGN performs much better than JODIE and DyRep, especially in terms of Positive accuracy and F1, with more than 10\% absolute improvement in average. This indicates that TGN is more effective at capturing temporal dependencies and evolving patterns of dynamic graphs. 

Consistent with the regression results, sampling more negative edges for training TGN can enhance the overall accuracy from 16.10\% to 50.37\%, but diminish the ability to classify positive edges, reducing it from 32.19\% to 12.91\%. Hence, selecting appropriate sampling methods is crucial to achieve a balance between positive and overall performance. 
\section{Limitation and Future Work}
We acknowledge several limitations of our work. Firstly, we did not conduct hyper-parameter searches to optimize the performance of the GNNs. Instead, we used default values for the hyper-parameters, which may not be optimal for all datasets and models. Future work can investigate the effect of different hyper-parameters on the performance of GNN models. Secondly, we did not explore enough model structures to compare their performance, such as temporal GNNs like TGAT \citep{xu2020inductive} and static GNNs like GIN \citep{xu2018how}. Additionally, we only evaluated the models on the UN Trade dataset due to the lack of publicly available dynamic graph datasets. Therefore, the generalizability of the GNN models to other dynamic graphs remains to be investigated in future work. Another limitation of the UN Trade dataset is that it only covers a period of around 30 years, which limits the number of timestamps available for training and evaluation. Future work could evaluate the effectiveness of the proposed methods with more timestamps to provide more comprehensive findings and analysis. It is worth noting that a concurrent work \cite{huang2023temporal} provides more relevant datasets. 

\section{Conclusion}
This work evaluates the performance of existing GNNs on edge regression tasks. We formulate the temporal edge regression task, predicting the actual edge weight or its magnitude in the graph. Three normalization methods are designed and applied to scale the original inputs for smooth training. We propose three simple but powerful baselines, which outperform most GNNs, indicating the research gap in leveraging GNNs for edge regression tasks. We apply both static and dynamic GNNs to this task to make a comprehensive evaluation. It shows that TGN outperforms the other GNN models, indicating its superiority in modeling temporal dynamics and dependencies of graph structures. Additionally, our analysis reveals that the proportion of negative edges in the training samples has a significant impact on test performance. Our work represents a step forward in modeling edge regression in dynamic graphs, and we call for future research to explore the potential of GNNs in addressing more complex temporal edge regression tasks.

\section*{Acknowledgements}

We would like to thank Prof. Pietro Liò and Dr. Petar Veličković for their kind support on this project. This study would not have been possible without their generous help.

\bibliographystyle{unsrtnat}
\bibliography{reference}

\appendix

\section{Dataset Information}
\label{a:data}

\begin{figure}[!ht]
    \centering
    \includegraphics[width=.6\textwidth]{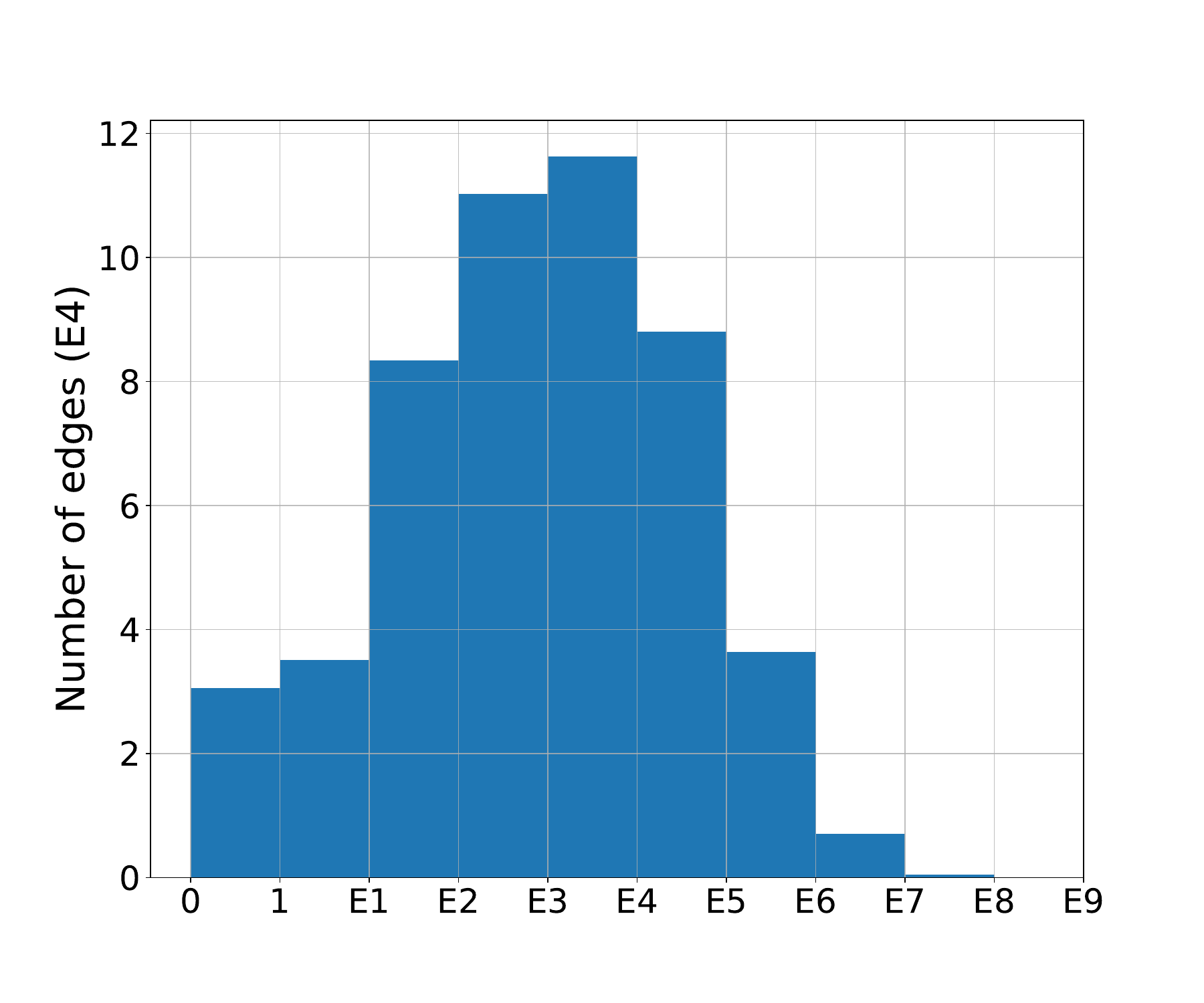}
    \caption{The distribution of edge weights in the UN Trade dataset. }
    \label{fig:distribution}
\end{figure}

The distribution of UN Trade dataset is illustrated in Figure~\ref{fig:distribution}. 

\section{Experimental Settings}
\label{experiment}
We use the default hyper-parameters in the original paper \cite{poursafaei2022towards}, which uses dynamic GNNs for link predictions in the UNTrade dataset. The Adam optimizer with a learning rate of $10^{-4}$ and a batch size of 200 is used. The number of epochs is set to 50, and an early stopping method is adopted with patience of 5. The dropout rate is configured to 0.1, and two attention heads are used. To eliminate randomness, we conduct all experiments with three random seeds to report the average values and standard deviations. Dynamic GNNs take approximately 30 minutes for each run on a single A100 GPU, while static GCN can finish one round in a minute on CPU.

\section{More Results}
\label{a1}

The experimental results including new nodes are reported below for the readers to have a more comprehensive understanding of GNNs on edge regression tasks. 

\begin{table}[!ht]
\caption{Loss with standard deviation (new nodes). Numbers in red mean the best results for all methods, and numbers in bold are the best results for GNNs.}
\centering
\resizebox{\textwidth}{!}{
\begin{tabular}{lcccccc}
\toprule
                           & \multicolumn{2}{c}{Log normalization}                                       & \multicolumn{2}{c}{\begin{tabular}[c]{@{}c@{}}Min-max normalization\\ $(10^{-2})$\end{tabular}} & \multicolumn{2}{c}{\begin{tabular}[c]{@{}c@{}}Node degree normalization\\ $(10^{-3})$ \end{tabular}} \\ \cmidrule(lr){2-3} \cmidrule(lr){4-5} \cmidrule(lr){6-7} 
                           & Positive MSE                 & Overall MSE                                      & Positive MSE                               & Overall MSE                                   & Positive MSE                                 & Overall MSE                                      \\ \midrule
\textbf{Baselines}         &                              &                                              &                                            &                                           &                                              &                                             \\
Mean                       & 2.785                        & 5.516                                        & 3.880                                       & 1.952                                     & 3.218                                        & 1.716                                       \\
Persistence forecast                 & {\color[HTML]{FF0000} 0.546} & 4.833                                        & {\color[HTML]{FF0000} 0.129}               & 1.690                                      & {\color[HTML]{FF0000} 1.402}                 & {\color[HTML]{FF0000} 1.069}                \\
Historical average        & 0.678                        & 4.344                                        & 1.167                                      & {\color[HTML]{FF0000} 1.167}              & 1.536                                        & 1.097                                       \\ \midrule
\textbf{Dynamic GNNs}      &                              &                                              &                                            &                                           &                                              &                                             \\
\multicolumn{7}{l}{\textbf{Train on all node pairs}}                                            \\
TGN                        & 9.043 (0.33)                 & 4.881 (0.01)                                 & 3.958 (0.00)                               & 1.980 (0.00)                              & 3.055 (0.00)                                 & 1.530 (0.00)                                \\
JODIE                      & 9.169 (0.07)                 & 5.146 (0.09)                                 & 37.19 (40.6)                               & 144.6 (105.)                              & 144.8 (128.)                                 & 117.2 (90.3)                                \\
DyRep                      & 9.488 (0.20)                 & 5.082 (0.07)                                 & 3.960 (0.00)                                & 1.981 (0.00)                              & 2.975 (0.01)                                 & 1.558 (0.02)                                \\ \midrule
\multicolumn{7}{l}{\textbf{Train on positive edges}}                                            \\
TGN                        & \textbf{1.834 (0.04)}        & 4.972 (0.42)                                 & \textbf{3.944 (0.00)}                      & 1.978 (0.00)                              & \textbf{2.867 (0.04)}                        & 1.522 (0.03)                                \\
JODIE                      & 3.059 (0.46)                 & 5.376 (0.04)                                 & 20.12 (22.7)                               & 31.21 (21.9)                              & 18.70 (12.7)                                 & 96.81 (2.90)                                \\
DyRep                      & 3.146 (0.20)                 & 4.067 (0.07)                                 & 3.945 (0.01)                               & 1.984 (0.01)                              & 3.193 (0.23)                                 & 1.929 (0.24)                                \\ \midrule
\multicolumn{7}{l}{\textbf{Train with negative sampling}}                                             \\
TGN                        & 3.593 (0.59)                 & {\color[HTML]{FF0000} \textbf{3.389 (0.07)}} & 3.947 (0.00)                               & \textbf{1.977 (0.00)}                     & 2.914 (0.04)                                 & \textbf{1.506 (0.03)}                       \\
JODIE                      & 5.369 (0.15)                 & 3.724 (0.03)                                 & 4.531 (0.64)                               & 2.803 (0.75)                              & 33.65 (11.5)                                 & 35.33 (13.4)                                \\
DyRep                      & 5.738 (0.30)                 & 3.898 (0.08)                                 & 4.154 (0.24)                               & 2.177 (0.24)                              & 3.731 (0.28)                                 & 2.663 (0.67)                                \\ \bottomrule
\end{tabular} }
\label{t2}
\end{table}
\begin{table}[!ht]
    \centering
    \caption{Accuracy and F1 score (\%) with standard deviation (new nodes). Numbers in red mean the best results for all methods, and numbers in bold are the best results for GNNs.}
    \begin{tabular}{lcccc}
\toprule
                           & Positive Accuracy            &  Overall Accuracy                                 & Positive F1                  &  Overall F1                        \\ \midrule
\textbf{Baselines}         &                              &                                              &                              &                              \\
Most                       & 22.73                        & 11.37                                        & 8.56                         & 2.37                         \\
Persistence forecast                 & {\color[HTML]{FF0000} 63.64} & 47.18                                        & {\color[HTML]{FF0000} 64.81} & {\color[HTML]{FF0000} 46.56} \\
 \midrule
\textbf{Dynamic GNNs}      &                              &                                              &                              &                              \\
\multicolumn{5}{l}{\textbf{Train on positive edges}}                              \\
TGN                        & \textbf{29.01 (1.04)}        & 14.50 (0.52)                                 & \textbf{26.66 (1.01)}        & 9.24 (0.28)                  \\
JODIE                      & 21.37 (1.45)                 & 10.69 (0.73)                                 & 13.88 (3.09)                 & 4.44 (1.27)                  \\
DyRep                      & 23.13 (1.69)                 & 11.57 (0.85)                                 & 17.05 (3.04)                 & 5.56 (1.18)                  \\ \midrule
\multicolumn{5}{l}{\textbf{Train with negative sampling}}                             \\
TGN                        & 8.16 (2.50)                  & {\color[HTML]{FF0000} \textbf{50.52 (0.71)}} & 10.41 (2.75)                 & \textbf{40.21 (2.06)}        \\ 
JODIE                      & 1.19 (0.02)                  & 50.09 (0.48)                                 & 1.06 (0.04)                  & 34.38 (0.26)                 \\
DyRep                      & 0.24 (0.3)                   & 50.00 (0.02)                                 & 0.14 (0.16)                  & 33.48 (0.15)                 \\ \bottomrule
\end{tabular}
\label{t4}
\end{table}

\end{document}